\documentclass[letterpaper, 10 pt, journal, twoside]{IEEEtran}

\usepackage[utf8]{inputenc}
\usepackage[T1]{fontenc}
\usepackage{graphicx}
\usepackage{amsfonts}
\usepackage{adjustbox}
\usepackage{amsmath} 
\usepackage{mathtools}
\usepackage{algorithm}
\usepackage{algpseudocode}
\usepackage{booktabs}
\usepackage{multirow}
\usepackage{amssymb}

\IEEEoverridecommandlockouts                              % This command is only needed if 
\usepackage{cite}
\usepackage{hyperref}
\usepackage[dvipsnames]{xcolor}

\newcommand{\HL}[1]{{\color{black}{#1}}}

\newcommand{\V}[1]{{\boldsymbol{\mathbf{#1}}}}

\usepackage[normalem]{ulem}

\title{Haptic Sorter: A Unified Planning Framework for Online Shape Estimation and Real-Time Pose Inference}
\author{Zhuoyi Lu$^{1,2}$, Lin Yang$^{2*}$, Sri Harsha Turlapati$^{2}$ and Domenico Campolo $^{2}$
\thanks{Manuscript received: May 29, 2026; Revised: August 27, 2026; Accepted: September 29, 2026.}
\thanks{This paper was recommended for publication by Editor Júlia Borràs Sol upon evaluation of the Associate Editor and Reviewers’ comments.}
\thanks{$^1$ is with Advanced Remanufacturing and Technology Centre (ARTC), Agency for Science, Technology and Research (A*STAR), Singapore}
\thanks{$^2$ are with the School of Mechanical and Aerospace Engineering, Nanyang Technological University (NTU), Singapore.}
\thanks{$^*$ Corresponding author: {\tt yang0752@e.ntu.edu.sg}}
\thanks{Digital Object Identifier (DOI): see top of this page.}
}
\date{}

\begin{document}

\markboth{IEEE ROBOTICS AND AUTOMATION LETTERS. PREPRINT VERSION. ACCEPTED SEPTEMBER, 2026}
{Lu \MakeLowercase{\textit{et al.}}: Haptic Sorter: A Unified Planning Framework for Online Shape Estimation and Real-Time Pose Inference}
\maketitle
% \thispagestyle{empty}
% \pagestyle{empty}

%%%%%%%%%%%%%%%%%%%%%%%%%%%%%%%%%%%%%%%%%%%%%%%%%%%%%%%%%%%%%%%%%%%%%%%%%%%%%%%%
\begin{abstract}
Robotics manipulation usually assumes that the shape and pose of the object are known to the robot prior to motion planning. However, precise geometric information is not always available in practice, and pose inference suffers from sensor uncertainties and view occlusion. In this work, we propose a unified model-based geometric framework integrating robotic haptic perception, modeling, and manipulation planning. Our novelties involve: \textit{i)} Introducing Bayesian Optimization (BO) to guide the haptic exploration for object shape inference, where superellipses are used to approximate geometric boundary; \textit{ii)} Adaptive formulation of manipulation potential encoding object geometry for quasi-static robot-object interaction; \textit{iii)} Proposing an online Ordinary Differential Equation (ODE) for real-time pose inference based on model prediction and tactile feedback. We validate the framework on three elementary geometries (circle, ellipse, rectangle) and a multi-superellipse case study (hexagon) in simulation and a real-world multi-arm setup for a 2D sorting task.
\end{abstract}
%%%%%%%%%%%%%%%%%%%%%%%%%%%%%%%%%%%%%%%%%%%%%%%%%%%%%%%%%%%%%%%%%%%%%%%%%%%%%%%%

%%%%%%%%%%%%%%%%%%%%%%%%%%%%%%%%%%%%%%%%%%%%%%%%%%%%%%%%%%%%%%%%%%%%%%%%%%%%%%%%
\section{INTRODUCTION}

\begin{figure}[!ht]
    \centering
    \includegraphics[width=0.95\linewidth, keepaspectratio]{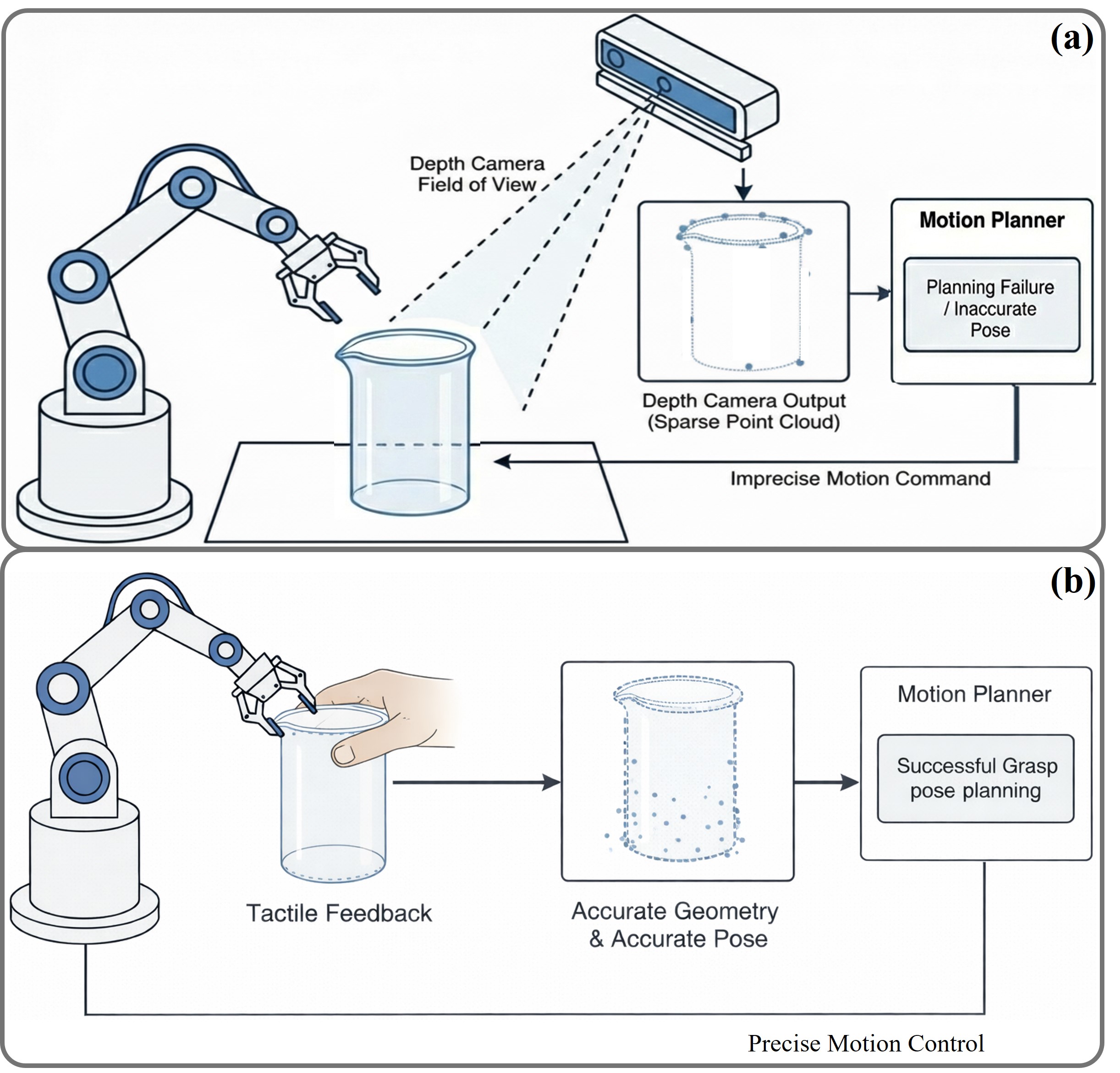}
    \caption{A motivating scenario of (a) A depth camera to capture the full geometry of a transparent glass beaker, resulting in failure of motion planner thus imprecise control command. (b) A successful motion planning trial with tactile feedback from touch.}
    \label{Fig1}
    \vspace{-5mm}
\end{figure}

Contact-rich manipulation requires accurate object geometry and pose. Collision-inclusive planners use exact meshes \cite{ren2025collision} or models such as superellipses (SQs) \cite{yang2025planning} to represent the object and robot.
These estimates must remain valid during execution, when contact-induced slip and pose drift alter the object state.
Vision can struggle with transparent or reflective objects, such as the glass beaker in Fig. \ref{Fig1}.
Furthermore, the robot end-effector may occlude the camera view, limiting precise perception during the contact phase \cite{schmidt2015dart}.

% \cite{Wang2025visual, dutta2025predictive}
For many years, scholars have been using vision-based approaches to perceive object geometry. \cite{dutta2025predictive} has developed a novel predictive perception framework for shape estimation by multiple camera views. The framework takes advantage of Bayesian Inference, while selecting the next camera views with maximum view entropy. The complete shape information can therefore be fitted by Superquadric Recovery with collected point cloud data. However, vision-based methods usually fall short in detecting object surfaces with low reflectivity,
% \LY{
and treat shape estimation as a one-shot pre-processing step loosely coupled to the downstream manipulation model.
% }
Extensive research investigates the visuo-tactile sensors for 3D surface identification. The sensor can provide high-fidelity local information through tactile feedback \cite{huang2025gelslam}. Nevertheless, high-resolution data makes the framework computationally demanding.

In parallel, haptics offers a complementary avenue to probe object pose and shape, like human who can locate and estimate object geometry and properties by leveraging the sense of touch \cite{luo2017robotic}. Gaussian Process (GP) is generally considered a favorable technique for robotic haptic exploration. It approximates object surfaces through sparse contact points, and the variance of GP naturally provides guidance by incorporating robotic travel costs \cite{ottenhaus2018active}. By formulating tactile inference as a SLAM problem, tactile feedback can be fused into a global model characterized by Gaussian process implicit surfaces (GPIS), while real-time estimating object pose with Factor Graph Formulation \cite{suresh2021tactile}. While these frameworks accumulate single-point contacts over time, we pursue a complementary strategy of simultaneous multi-contact measurements that provides instantaneous observability.
Additionally, the choice of the GP covariance function is crucial, as it determines how GP smooths out outlier measurements for shapes with sharp edges. A recent work encodes symmetry into the GP covariance for 3D active exploration \cite{bonzini2025robotic}, at the cost of more contact budget and expensive runtime evaluation. Extending pose estimate from $\mathrm SE(2)$ to $ \mathrm SE(3)$ also warrants dedicated treatment \cite{chen2026wrench}. More fundamentally, these approaches treat haptic estimation as an isolated perception problem: 
% \LY{
the recovered surface is represented by the full set of contact observations, making it expensive to embed into an online gradient-based manipulation model.
% }

To mitigate the limitations of previous methodologies, our work introduces an online closed-loop planning framework incorporating haptic exploration and manipulation planning in a single differentiable model. We employ a SQ-based object representation to naturally capture the property of symmetry. 
% \LY{
Its analytic form lets the recovered shape directly instantiate a differentiable manipulation potential, so that the same model both predicts contact forces and, via its gradient, drives an online pose update.
% } 
Intuitively, we guide the robotic haptic exploration process by Bayesian Optimization (BO), which motivates the robot to explore the regions with the largest radius (e.g., corners) by maximizing the Expected Improvement (EI) as the acquisition function \cite{zhou2024corrected}. This favors SQ recovery \cite{liu2022robust} by prioritizing the geometric features that most strongly constrain the fit, while preserving full differentiability of the recovered shape.

% Adaptive ODE + Haptic ODE
% \LY{
This differentiability allows the recovered shape to be embedded into a potential-based manipulation framework. Building on the quasi-static formulation of
% }
\cite{campolo2025geometric}, which splits variables into robot internal states $\mathbf{z}$ and controllable DoF $\mathbf{u}$ along an implicitly defined equilibrium manifold ($\mathcal{M}^{eq}$). We extend this framework by defining a manipulation potential $\mathcal{W}(\mathbf{z},\mathbf{u}, \mathbf{\zeta})$ 
% \LY{
instantiated directly from the recovered SQ parameters. Given current states $\{\mathbf{z}_0,\mathbf{u}_0 ,\mathbf{\zeta}_0\}$ and intended robot actions {$\mathbf{u}_1$}, the system evolution and interaction forces are propagated through an Adaptive ODE along $\mathcal{M}^{eq}$. In parallel, an online Haptic ODE uses the residual between sensed and model predicted forces as a feedback signal for online pose estimation. Together, the two ODEs close the loop on online pose estimation and manipulation planning.
% }

% Scope, objective and Significance
% \LY{
A 2D sorting task is validated because it concentrates the failure modes a model-based observer must handle: imperfect grasping, slip, and pose drift during contact. We implement the framework on a multi-arm setup where a pair of planar 2-DoF arms with F/T sensors provides two haptic measurements, while a 6-DoF manipulator executes the sorting trajectory under continuous correction from the Haptic ODE. 
% \LY{
Across exploration, modeling, and tracking, the same SQ parameterization serves as the shared representation.
% } 
Our contributions are:
% }
\begin{enumerate}
    \item A BO-guided haptic exploration strategy that recovers superellipse parameters from sparse tactile contacts;
    \item An adaptive manipulation potential formulation based on recovered superellipse geometry, modeling the quasi-static interaction between robots and objects;
    \item A novel Haptic ODE that utilizes the residual between modeled forces and tactile feedback for online pose estimation, enabling closed-loop correction under physical uncertainties.
\end{enumerate}

\begin{figure}[htbp!]
    \centering
    \includegraphics[width=0.95\linewidth, keepaspectratio]{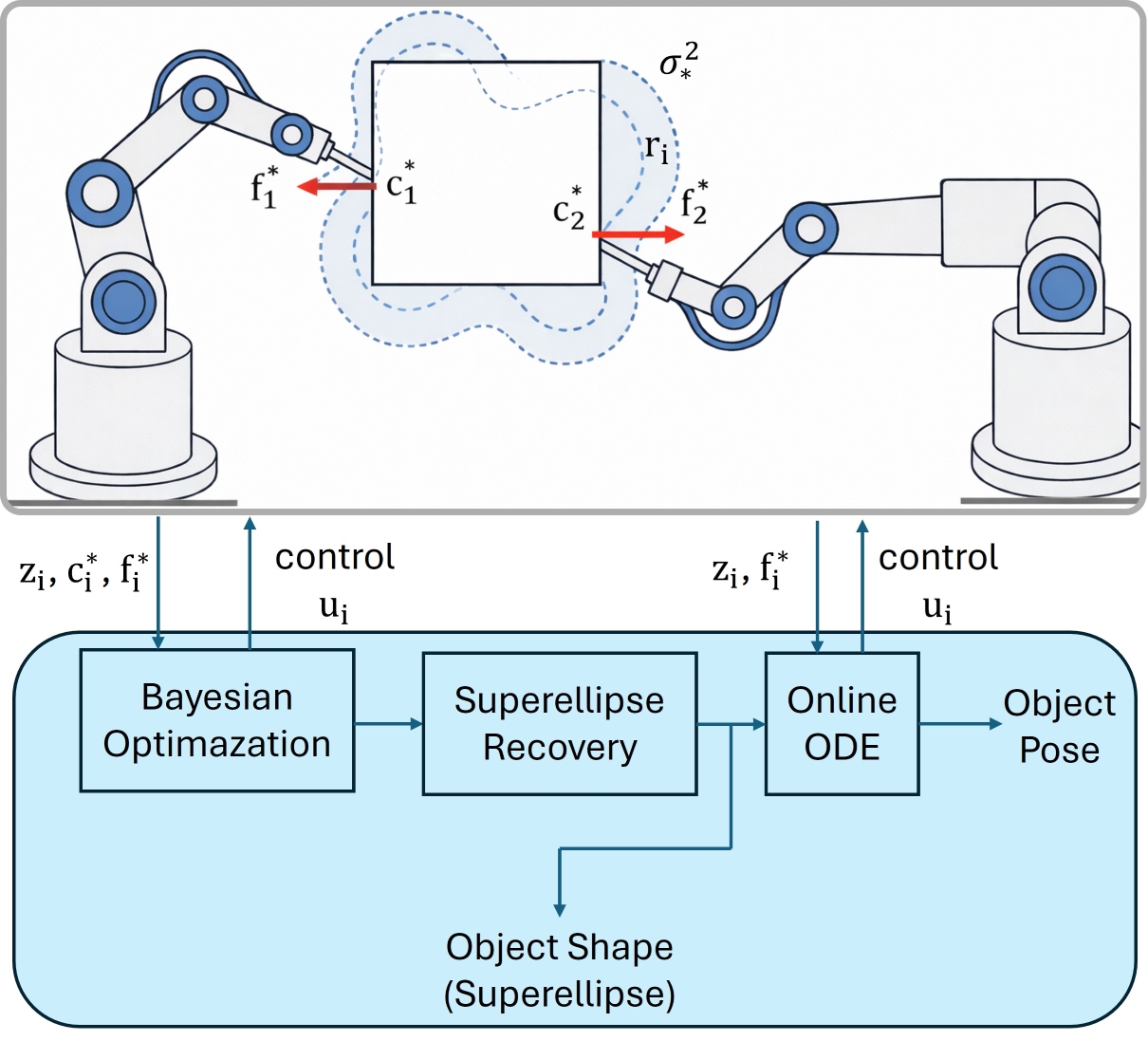}
    \caption{Haptic Exploration with dual 2-DoF planar arms for shape and pose inference, starting with a weak prior pose from camera and held by a 6-DoF manipulator on a table.}
    \label{Fig2}
    \vspace{-3mm}
\end{figure}

\section{Problem Formulation}
In this work, we investigate the problem of object 2D shape and pose inference through robotic haptic exploration by a pair of 2-DoF planar dual arms. As depicted in Fig.~\ref{Fig2}, the dual arms begin exploration given a weakly initialized pose from the camera, while a 6-DoF manipulator holds the object in place throughout exploration.
\\
\textbf{Object Pose}: \HL{We define the object pose at the current timestamp t as $\V \zeta(t) = (x(t), y(t), \theta(t)) \in \mathrm{se}(2)$ and the corresponding homogeneous  transformation matrix $T(\boldsymbol{\zeta}(t)) = 
\begin{bmatrix} 
R(\theta(t)) & \mathbf{t}(t) \\ 
\mathbf{0}^\top & 1 
\end{bmatrix} \in \mathrm{SE}(2)$.}
\\
\textbf{Object Shape}: The object boundary is approximated 
by superellipses, described by the parameter set $\mathcal{S} = \{a_1, a_2, \epsilon\}$, where $a_1, a_2$ 
scales semi-axes and $\epsilon$ is the shape exponent.
\\
\textbf{Contact}: We define haptic measurements to be $\mathbf{f_i}^* = [f_{x}(t), f_{y}(t)]^T \in{\mathbb{R}^ 2}$ with end-effector states $\mathbf{z}_i = \begin{bmatrix}
    z_{x}, z_{y}
\end{bmatrix}^T\in\mathbb{R}^2$. Contact points $\mathbf{c}^*$ are computed via wrench axis from the knowledge of $\mathbf{f}^*_i$, $\mathbf{z}_i$, and probe radius $r_{p}$ \cite{turlapati2022towards}. 
\\
For simplicity, we make several \textbf{assumptions}:
\begin{enumerate}
    \item Quasi-static interaction simplifies the problem by focusing on contact forces while neglecting inertia and Coriolis effects \cite{yang2025planning}.
    \item The geometric shape is well-approximated by the superellipse family for a compact analytic form.
    \item The rough scale and pose $\hat \zeta_{cam}$ are given by a camera.
\end{enumerate}
The accumulated contact points $\{\mathbf{c}^*\}$ are subsequently fed into Superellipse Recovery 
\cite{liu2022robust}, which regresses the shape parameters $\{a_1, a_2, \epsilon\}$. These parameters directly instantiate the manipulation potential and initiate robotic sorting.

Building upon the haptic exploration phase, the robotic sorting task integrates real-time pose estimation with trajectory re-planning to accurately manipulate the object from an initial pose $\zeta_0 = \{x_0,y_0,\theta_0 \}$ to a desired target pose $\zeta_{goal} = \{x_g,y_g,\theta_g \}$. As detailed in Algorithm \ref{alg:haptic_sorting}, the dual planar arms operate under impedance control to compliantly guide the object, while the 6-DoF manipulator executes the primary trajectory using position control based on continuous updates from an asynchronous observer (Haptic ODE). To counteract inevitable physical disturbances (i.e., gripper misalignment, grasp slips), the system periodically updates the estimated object pose $\hat \zeta$ and generates  trajectories using Dynamical Movement Primitives (DMPs) \cite{ijspeert2013dynamical}.

\begin{algorithm}[H]
\caption{Sorting with Haptics: Trajectory Execution and Re-planning}
\label{alg:haptic_sorting}
\begin{algorithmic}[1]
\Procedure{SortWithHaptics}{$\zeta_{goal}, \hat\zeta, \mathbf{z}^* = \mathbf{z}_0$}
\State $Init(Haptic ODE())$
\Comment{ODE Initialization}
\While{$|| \hat\zeta - \zeta_{goal} ||_\Sigma \geq \zeta_{thresh}$}
\State $DMP() \leftarrow \zeta_{goal}, \hat\zeta$
\State $T_{end}, u_{dual}, u_{6DOF} \leftarrow Solve(DMP())$
\While{$t < T_{end}$}
\State $ImpeCtrl(u_{dual}(t))$ 
\Comment{Dual Arm}
\State $PosCtrl(u_{6DOF}(t))$ 
\Comment{6-DoF Arm}
\State $\mathbf{f}_{sen} \leftarrow ReadDualForce()$
\If{$Haptic ODE().done$}
\State $\hat\zeta, \mathbf{z}^* \leftarrow Haptic ODE()$
\State $AsySolve(Haptic ODE()) \leftarrow \hat\zeta, z^*, \mathbf{f}_{sen}$
\EndIf

\EndWhile

\EndWhile
\State $openGripper()$
\Comment{End Task Upon Convergence}
\EndProcedure
\end{algorithmic}
\end{algorithm}

\vspace{-4mm}
\section{Methodology: Bayesian Optimization, Superellipses and Proxies}

\subsection{Bayesian Optimization}

In the robotic haptic exploration, we represent the object surface as a continuous mapping $g: [0, 2\pi) \to \mathbb{R}$. We model this unknown surface as a Gaussian Process (GP), such that $g(\theta) \sim \mathcal{GP}(m(\theta), k(\theta, \theta'))$, where $m(\theta)$ is the mean function and $k(\theta, \theta')$ denotes the covariance kernel. Given a set of discrete observations $\mathcal{D} = \{ (\theta_i,r_i)\}^{i\in[1,n]}$, the observed radial distances are defined as $r_i = g(\theta_i) + \epsilon$, where $\epsilon \sim \mathcal{N}(0, \sigma_n^2)$ accounts for the Gaussian noise.

To ensure the regressed shape is 
continuous while remaining capable of capturing high-curvature features (e.g., vertices of a superellipse), we employ a Matérn kernel ($v = 5/2$) as the surrogate model \cite{lim2021extrapolative}. Furthermore, to enforce a periodicity of $2\pi$ and ensure geometric closure, we apply a unit-circle mapping to the angular inputs. This transforms the domain of the mapping 
into:
\begin{equation}
\label{GPmapping_circle}
F: (\cos\theta, \sin\theta) \mapsto r, \quad \theta \in [0, 2\pi)
\end{equation}

For a test point $\theta_*$, the GP yields a Bayesian posterior distribution $F_* \sim \mathcal{GP}(\bar{F_*},\sigma_*^2)$, which characterizes the predictive radius and associated model uncertainty. The posterior mean and variance are computed as \cite{rasmussen2003gaussian}:
\begin{equation}
\label{GPoutput}
\begin{split}
\bar{r}_* &= k_*^T(\mathbf{K} + \sigma_{\mathrm{noise}}^2 \mathbf{I}\big)^{-1} \mathbf{Y}
\\
\sigma_*^2 &= k_{**} - k_*^T(\mathbf{K} + \sigma_{\mathrm{noise}}^2 \mathbf{I}\big)^{-1} k_*
\end{split}
\end{equation}
where $\mathbf{K} \in \mathbb{R}^{n \times n}$, $k_* \in \mathbb{R}^{n \times 1}$ and $k_{**} \in \mathbb{R}$ are the train-train, train-test, and test-test kernels, respectively.

\begin{figure}[htbp!]
    \centering
    \includegraphics[width=\linewidth, keepaspectratio]{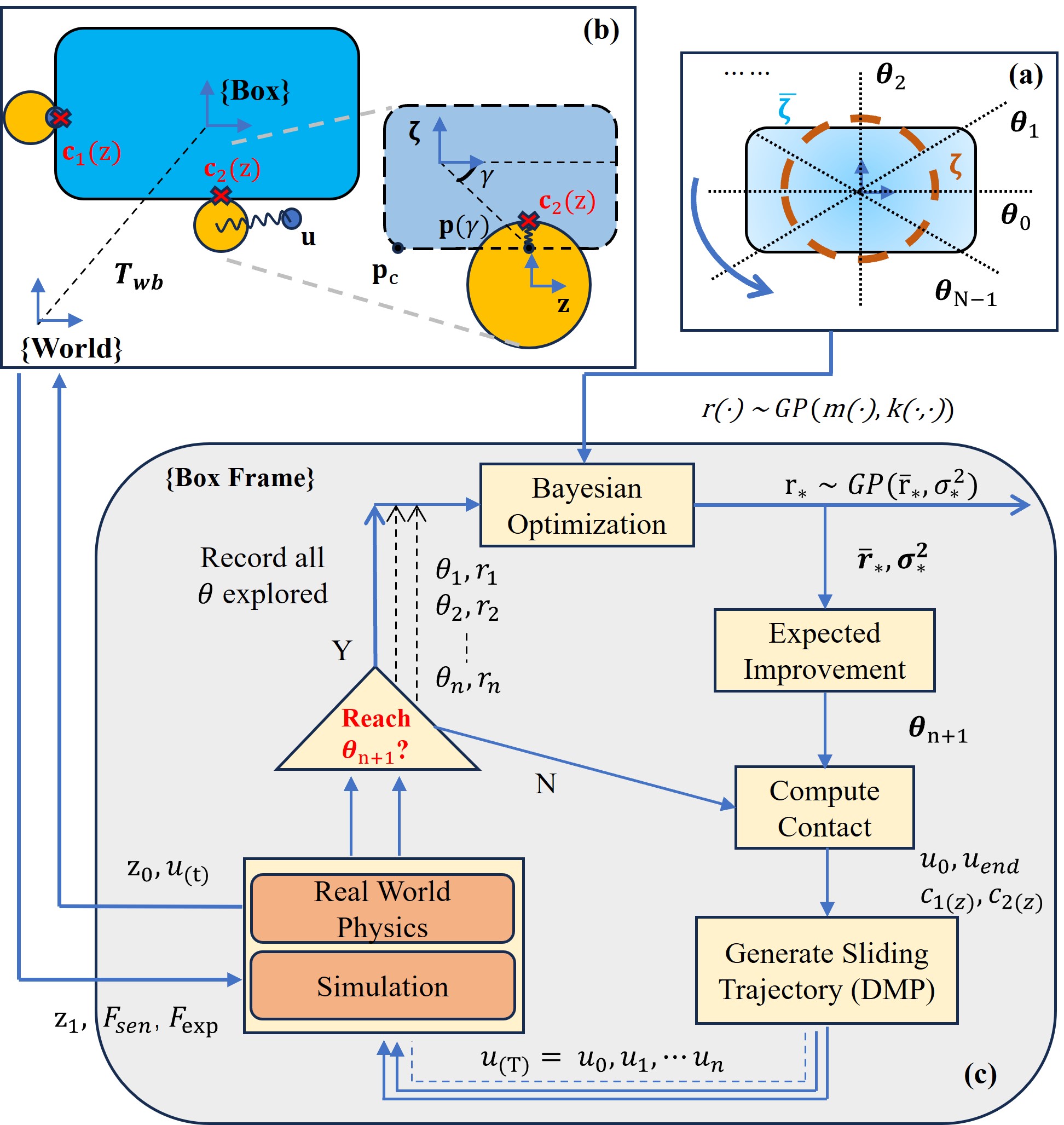}
    \caption{A Conceptual Workflow of BO Algorithm: (a) Initial configuration; (b) System modeling of two hand touching a box. Contacts and forces are measured in the world frame $\{\mathcal W\}$, while geometric properties (i.e. proxies, normals) are calculated in the object frame $\{\mathcal B\}$. We denote $T_{\mathcal{WB}}\in \mathrm SE(2)$ as the homogeneous transformation between two frames. (c) Block diagram of BO algorithm to guide robotic haptic exploration. Here $\mathbf{u}{(T)}$ denotes the discrete way-points generated via DMP and $\mathbf{u}{(t)}$ is the time-interpolated action.}
    \label{BO}
    \vspace{-2mm}
\end{figure}

% Then discuss the BO
The exploration starts with a rough object pose $\mathbf{\zeta}$ perceived from the camera, aside from the ground truth pose $\mathbf{\bar\zeta}$, as well as a weak circular shape prior (brown dotted circle) for GP. As shown in Fig.\ref{BO}(c), the BO algorithm aims to find the angle segment maximizing the learned radius:
$
   \max_{\theta_i \in [0,2\pi]} g(\theta) 
$.

% Start to introduce acquisition function 
To balance exploration and exploitation, Expected Improvement (EI) is selected as the acquisition function, with a closed form expression \cite{garnett2023bayesian}:
\begin{equation}
\mathrm{EI}_n(\theta) = \Delta_n(\theta)\Phi\left(\frac{\Delta_n(\theta)}{\sigma_n(\theta)}\right) + \sigma_n(\theta)\phi\left(\frac{\Delta_n(\theta)}{\sigma_n(\theta)}\right)
\label{eq:EI}
\end{equation}
where $\Delta_n(\theta) := \mu_n(\theta) - r_n^*$
is the expected difference in quality between the proposed point \(\theta\) and the currently largest observed radius \(r_n^*\). Here, \(\phi(\cdot)\) and \(\Phi(\cdot)\) denote the Probability Density Function (PDF) and Cumulative Distribution Function (CDF) of the standard normal distribution, respectively, and $\sigma_n(\theta)$ represents the posterior predictive standard deviation at $\theta$.
The BO  evaluates the objective at the point with the largest expected improvement as $\theta_{n+1} = \arg\max_\theta\mathrm{EI}_n(\theta)$. Furthermore, evaluating this objective globally over a discrete observation space prevents the system from being trapped in local maxima.

% Lastly how we haptic plays its role
 In each BO iteration illustrated in Fig. 3(c), the robotic probe closest to the target angle $\theta_{n+1}$ is commanded to slide along the object surface. The algorithm evaluates the radius by commanding the robot to move to $\theta_{n+1}$, where its motion is governed by a sliding trajectory.
\begin{equation}
    \mathbf{u}_0 = \mathbf{z}_{i,0} \;,
\quad
    \mathbf{u}_T = \mathbf{z}_{i,0} + \alpha_{s}\mathbf{R}\mathbf{\hat{f}_{sen,i}}  
\end{equation}
where $\mathbf{\hat{f}_{sen,i}}$ is the normalized force vector sensed by the robot, $\mathbf{R}\in\mathbb{R}^{2\times2}$ is the orthogonal rotation matrix, and $\alpha_{s}$ scales how far the robotic probe slides. 

During execution, acquired contact points augment the training set to update the radius prediction $r_*$ and associated uncertainty over the domain $\theta_* \in[0,2\pi]$ with output mean and variance (Eq. \ref{GPoutput}), while the next angle segment is evaluated by maximizing Eq.\ref{eq:EI}. The algorithm terminates when the maximum predictive variance falls below a predefined threshold $\sigma_{thresh}$, indicating its convergence. The accumulated contact points are subsequently fed into the Superellipse Recovery \cite{liu2022robust}. The recovered shape parameters $\{a_1, a_2, \epsilon\}$ directly instantiate the manipulation potential ${W}(\mathbf{z}, \mathbf{u}, \V \zeta)$.

\subsection{Quasi-static Modeling of the Mechanical System}
Highlighted in \cite{campolo2023quasi}, the mechanical robot-environment interaction under quasi-static assumption can be treated as a control problem with directly controlled variable $\mathcal{U}$ and non-directly controlled variable $\mathcal{Z}$. The total configuration space can be defined as the product manifold 
$Q = \mathcal{Z\times U \times \zeta} \subset{\mathbb{R}^M} \times \mathbb{R}^K \times \mathbb{R}^O$, 
where $\mathcal{Z}$ denotes the $M$-dimensional manifold of internal states, $\mathcal{U}$ stands for the $K$-dimensional space of controllable degrees of freedom (DoFs) and $\mathcal{\zeta}$ indicates all the possible object pose configurations.
The Manipulation Potential $W(\mathbf{z},\mathbf{u},\V \zeta)$ of the system can be defined as a smooth field on the space $W : Q = \mathcal{Z\times U \times \zeta} \mapsto \mathbb{R}$. Since the system is always required to be in force equilibrium internally (i.e., only influenced by conservative forces like gravitational and spring energy), this poses a constraint on equilibrium state $\mathbf{z}^*(\mathbf{u}, \V{\zeta})$:
\HL{
\begin{equation}
\label{quasi-static constrain}
    \partial_\mathbf{z} W(\mathbf{z}^*(\mathbf{u}, \V{\zeta}),\mathbf{u},\V \zeta) = 0 \in \mathbb{R}^M
\end{equation}
}
A point is strictly stable when its Hessian is positive definite, i.e., $\partial_\mathbf{zz}^2 W|_{*} \succ 0$. Assuming the Hessian $\partial_\mathbf{zz}^2 W \in \mathbb{R}^{M \times M}$ is of full rank when \HL{$\partial_\mathbf{z} W(\mathbf{z}^*(\mathbf{u}, \V{\zeta}),\mathbf{u},\V \zeta) = 0$}, the implicit function theorem yields a locally unique map $\mathbf z^*(\mathbf u,\V \zeta)$ \cite{spivak2018calculus}. Subsequently, an equilibrium manifold $\mathcal{M}^{eq}$ can be defined as the smooth embedded sub-manifold within the ambient space $Q = \mathcal{Z\times U \times \zeta}$, subjected to the constraint (Eq. \ref{quasi-static constrain}):
\HL{
\begin{equation}
    \mathcal{M}^{eq} \coloneqq \{ (\mathbf{z},\mathbf{u},\V \zeta) \in \mathcal{Z\times U \times \zeta} | \partial_z W(\mathbf{z},\mathbf{u},\V \zeta) = 0\} 
\end{equation}
}
Under the quasi-static assumption, the total external force exerting on the robot should be equal to the control force. Intuitively, we can delineate properties regarding the robot-environment interaction, like the expected force $\mathbf{f}_{exp}$:
\HL{
\begin{equation}
\label{ctrl_force}
    \mathbf{f}_{exp} = -\partial_\mathbf{u} W(\mathbf{z}^*(\mathbf{u}, \V{\zeta}),\mathbf{u},\V \zeta)
\end{equation}
}

\subsection{Superellipses and Contact Modeling}
Leveraging the superellipse parameterization derived from haptic exploration, we apply our quasi-static framework to model the system via a differentiable manipulation potential, with following key components:

\subsubsection{Superellipses}
Superellipses can be defined implicitly with an \textbf{inside-outside} function $F(x,y)$ \cite{gridgeman1970lame}:
\HL{
\begin{equation}
\label{eq:inside-outside function}
    F(x,y) = |\frac{x}{a_1}|^\frac{2}{\epsilon} + 
            |\frac{y}{a_2}|^\frac{2}{\epsilon} - 1
\end{equation}
}
Here, $\epsilon$ as the exponent characterizes the shape of the superellipse, while $a_1, a_2$ controls its size. 

For a point $(x, y) \in \mathbb{R}^2$, whenever $F(x,y) > 0$ the point is outside of the superellipse. If $F(x,y) = 0$, the point is lying on the boundary of the superellipse and it is inside if $F(x,y) < 0$. Upon on Eq.\ref{eq:inside-outside function}, an angular parametrization of $\gamma$ can be applied to superellipses:
\HL{
\begin{equation}
\label{proxyDef}
\begin{aligned}
    \mathbf{p}(\gamma) &= R_{\mathcal{WB}}(\V \zeta) \ \mathbf{r}(\gamma)
        \begin{bmatrix} \cos\gamma \\ \sin\gamma \end{bmatrix} + \mathbf{t}_{\mathcal{WB}}(\V \zeta),
        \quad 0 \le \gamma \le 2\pi \\
    \mathbf{r}(\gamma) &= \left( \left|\frac{\cos\gamma}{a_1}\right|^{\frac{2}{\epsilon}}
        + \left|\frac{\sin\gamma}{a_2}\right|^{\frac{2}{\epsilon}} \right)^{-\frac{\epsilon}{2}}
\end{aligned}
\end{equation}
Here, proxies $\mathbf{p}(\gamma)$ are computed in $\mathcal{B}$ and transformed to $\mathcal{W}$ via $T_{\mathcal{WB}}(\V \zeta)$. We apply the same transformation for surface normals $\mathbf n^* = R_{\mathcal{WB}}(\V \zeta) \frac{\nabla F(\mathbf c_{ij}^{B*})}{||\nabla F(\mathbf c_{ij}^{B*})||}$.
}
\subsubsection{Contact Modeling}
\begin{figure}[!ht]
    \centering
    \includegraphics[width=\linewidth, keepaspectratio]{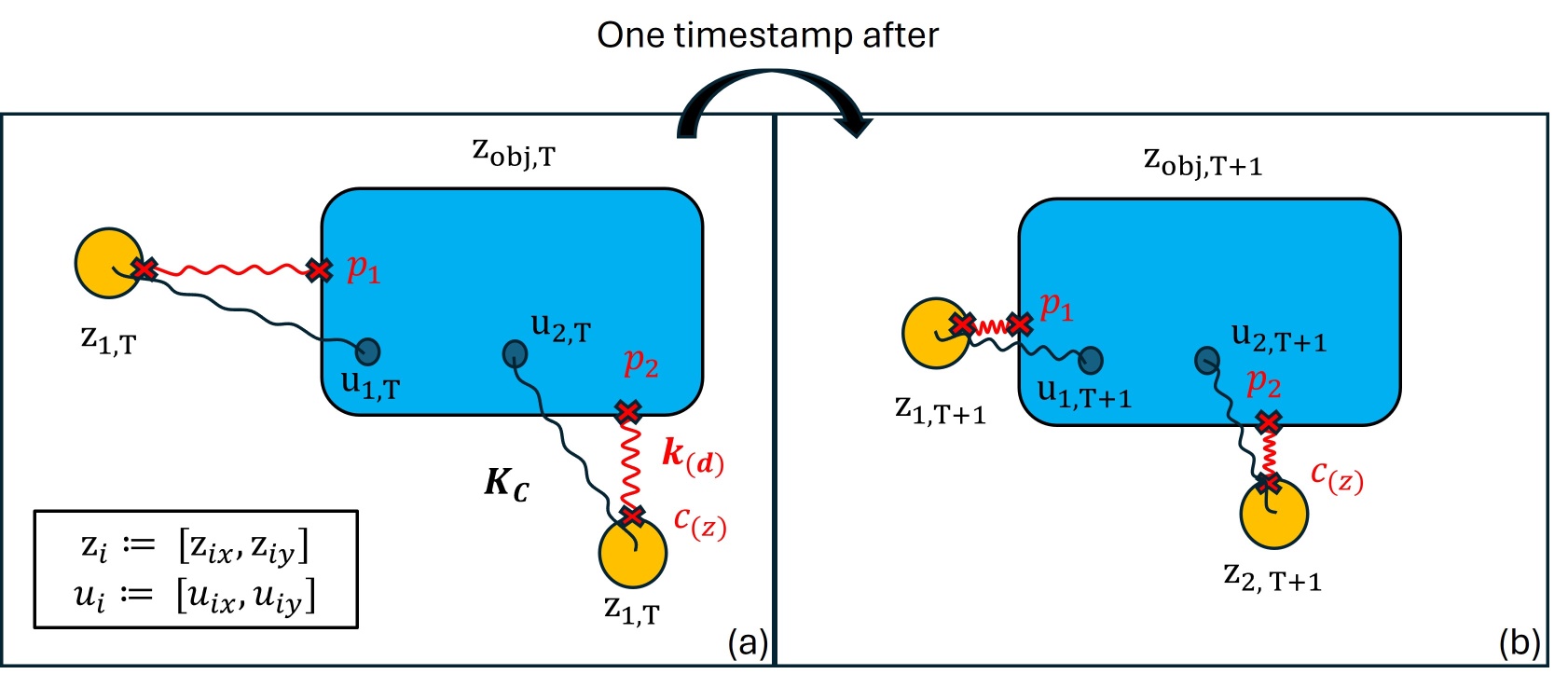}
    \caption{(a) A detail sketch of two robotic probe moving toward a rectangular box. (b) System evolution after one timestamp, showing the iteration of robot state $\mathbf{z}$ and proxies $\mathbf{p(\gamma)}$.}
    \label{fig:system model}
\end{figure}

Define a contact point $\mathbf{c(z_i)}$ on the geometric surface of robotic end-effector, there is a corresponding proxy $\mathbf{p^B}(\gamma) \in \mathbb{R}^2$ on the object surface denoting the closest point to the probe \cite{kana2021human}. The problem can be formulated into a minimization problem:
\begin{equation}
\label{proxySolve}
    \arg\min_{\gamma} || \mathbf{c(z_i)} - \mathbf{p(\gamma)}||, 
    \quad \text{s.t.} \quad \gamma\in[0,2\pi]
\end{equation}
\HL{To characterize the contact interaction between the robotic probe and the object, we adopt a non-linear spring with  stiffness $k(d)$ from the relaxed-Signorini family of contact models, where $d=F(\mathbf c^{\mathcal B})$, with $\mathbf c^{\mathcal B}=R_{\mathcal{WB}}^\top(\mathbf c-\mathbf t_{\mathcal{WB}})$ in the object frame.} The non-linear stiffness is modeled as:
\begin{equation}
    k(d) = k_{min} + \frac{1-\tanh{(d/d_0)}}{2} k_{max}
\end{equation}
where $d_0$, $k_{min}$, and $k_{max}$ regulate the behavior of the non-linear spring. \HL{This ensures differentiability and a smooth, continuous transition between contact and non-contact modes\cite{le2024contact}.}
\HL{The variable $d$ is the inside-outside level-set value computed from Eq.\ref{eq:inside-outside function}.} When $d\leq0$, the spring stiffness increases rapidly, modeling contact interactions. In contrast, the contact force is negligible when $d>0$. As we solve Eq. \ref{proxySolve} iteratively, proxies remain to trace the closest point on the object to their respective robot agents as shown in Fig.\ref{fig:system model}.

% \vspace{-2mm}

%
\subsection{System Modeling with Manipulation Potential}
As illustrated in Fig.\ref{fig:system model}, two circular robot probes $\mathbf{z}_i$ interact with the object $\V\zeta$ via the control input
$\mathbf{u}_i = \begin{bmatrix}
    u_{ix}, u_{iy}
\end{bmatrix}^T \in \mathbb{R}^2$ 
in the 2D task frame. The impedance controller implements a virtual force $\mathbf{f}_i\in\mathbb{R}^2$ on the object through a constant diagonal stiffness matrix $\mathbf{K}\in\mathbb{R}^{2\times2}$, as determined by Eq.\ref{ctrl_force}.
Furthermore, two proxies are introduced to track the contact points between the robotic probes and the object's superellipsoid boundary $\mathbf{p}_{ij}(\gamma_j,\V {\zeta})$. Here, 
$i$ and $j$ represent the indices for the respective robot probe and the object, ensuring each robot maintains a unique proxy.
Consequently, we can formulate the manipulation potential \HL{$W(\mathbf{z}^*(\mathbf{u}, \V{\zeta}),\mathbf{u},\V\zeta)$} with the configuration defined as 
\HL{$\V{z} = [\mathbf{z}_1, \mathbf{z}_2, \gamma_1, \gamma_2]^T$:
% proxy p is a function of zeta
% \begin{equation}
% \begin{split}
%     W(\mathbf{z}^*(\mathbf{u}),\mathbf{u}, {\V \zeta}) &= W_{ctrl} + W_{contact} \\
%               &= \sum_i \frac{1}{2} (\mathbf{u}_i - \mathbf{z}_i)^\top \mathbf{K_c} (\mathbf{u}_i - \mathbf{z}_i) \\
%               &+ \sum_i \sum_j \frac{1}{2} k(d_{ij}) \| \mathbf{c}_{ij}(\mathbf{z}_i) - \mathbf{p}_{ij}(\gamma_j, \V \zeta) \|^2
% \end{split}    
% \end{equation}
\vspace{-1mm}
\begin{align}
    W(\mathbf{z}^*(\mathbf{u}, \V{\zeta}),\mathbf{u}, {\boldsymbol{\zeta}}) &= W_{\text{ctrl}} + W_{\text{contact}} \label{eq:my_label} \\
              &= \sum_i \frac{1}{2} (\mathbf{u}_i - \mathbf{z}_i)^\top \mathbf{K}_c (\mathbf{u}_i - \mathbf{z}_i)+  \notag \\
              &\quad \sum_i \sum_j \frac{1}{2} k(d_{ij}) \| \mathbf{c}_{ij}(\mathbf{z}_i) - \mathbf{p}_{ij}(\gamma_j, \boldsymbol{\zeta}) \|^2 \notag
\end{align}
\vspace{-1mm}
}

The manipulation potential is constituted by two main components: $W_{ctrl}$ addresses the spring potential energy for the impedance controller; $W_{contact}$ signifies the contact interaction between the manipulated object and the robotic probe due to proxies.

\subsection{Adaptive ODE and Haptic ODE}
To explore $\mathcal{M}^{eq}$, we adapt a fundamental yet important result from a book insertion case study on quasi-static manipulation, which employs an adaptive Ordinary Differential Equation (ODE) approach \cite{yang2025planning}:
\begin{equation}
\label{adaptiveODE}
    \mathbf{\dot{z}} = -(\partial_\mathbf{zz}^2 W)^{-1} \partial_\mathbf{uz}^2 W \mathbf{\dot{u}}
              - \eta (\partial_\mathbf{zz}^2 W)^{-1} \partial_\mathbf{z} W
\end{equation}
Eq. \ref{adaptiveODE} governs the evolution of the system via two components: the former one incorporates the linear relationship between the infinitesimal changes in $\mathbf{z}$ and $\mathbf{u}$, while the second term provides a Newton-Raphson correction to prevent numerical drift from the equilibrium manifold.

%% version LY
Let $\V{x}=[x,y]^\top$ denote the translational component of the object pose.
We define the translational and rotational residual objectives as
\vspace{-1mm}
\begin{equation}
\begin{aligned}
J_T
&=
\frac{1}{2}\textstyle\sum_{i=1}^{2}
\left[
\V{n}_i^{*\top}
(\V{f}_{sen,i}-\V{f}_{exp,i})
\right]^2,
\\[-1mm]
-\frac{\partial J_T}{\partial \V{x}}
&\approx
\textstyle\sum_{i=1}^{2}
k_{eff,i}
\left[
\V{n}_i^{*\top}
(\V{f}_{sen,i}-\V{f}_{exp,i})
\right]\V{n}_i^*,
\\[-1mm]
J_R
&=
\textstyle\sum_{i=1}^{2}
\left(
1-\hat{\V{f}}_{sen,i}^{\top}\hat{\V{f}}_{exp,i}
\right),
\\[-1mm]
-\frac{\partial J_R}{\partial\theta}
&=
\textstyle\sum_{i=1}^{2}
\left(
\hat{\V{f}}_{exp,i}\times\hat{\V{f}}_{sen,i}
\right).
\end{aligned}
\end{equation}
\vspace{-1mm}

Here, $\hat{\V{f}}=\V{f}/\|\V{f}\|$, $\V{n}_i^*$ is the outward surface normal at the predicted contact, and $k_{eff,i}>0$ denotes the local effective stiffness relating normal penetration to the predicted normal force. These gradients are obtained under a local fixed-contact/proxy approximation. Using the force magnitudes as positive contact-dependent weights gives the Haptic ODE
\vspace{-1mm}
\begin{equation}
\dot{\zeta}=
\begin{bmatrix}
\alpha_T\textstyle\sum_{i=1}^{2}
(\V{f}_{sen,i}-\V{f}_{exp,i})^{T}\V{n}_i^*
\cdot\V{n}_i^*
\\[-0.5mm]
\alpha_R\textstyle\sum_{i=1}^{2}
(\V{f}_{exp,i}\times\V{f}_{sen,i})
\end{bmatrix}.
\end{equation}
\vspace{-1mm}

For local convergence, let
$\Delta\V{x}=\V{x}-\bar{\V{x}}$ and
$\Delta\theta=\theta-\bar{\theta}$. Linearization gives
\vspace{-1mm}
\begin{equation}
\dot{\Delta\V{x}}
\approx-\alpha_TA_T\Delta\V{x},
\qquad
\dot{\Delta\theta}
\approx-\alpha_RK_R\Delta\theta.
\end{equation}
\vspace{-3mm}

Here,
$A_T=\textstyle\sum_{i=1}^{2}
k_{eff,i}\V{n}_i^*\V{n}_i^{*\top}$
and
$K_R=\textstyle\sum_{i=1}^{2}
\|\V{f}_{exp,i}\|\|\V{f}_{sen,i}\|$.
Translation is locally exponentially convergent when $A_T\succ0$, which requires the contact normals to span $\mathbb{R}^2$; collinear normals are unobservable, while near-collinear normals slow convergence. Rotation is locally convergent for rotationally informative contacts, whereas rotationally symmetric objects such as a circle have unobservable absolute orientation.

\vspace{-3mm}
\begin{figure}[H]
    \includegraphics[width=\linewidth, keepaspectratio]{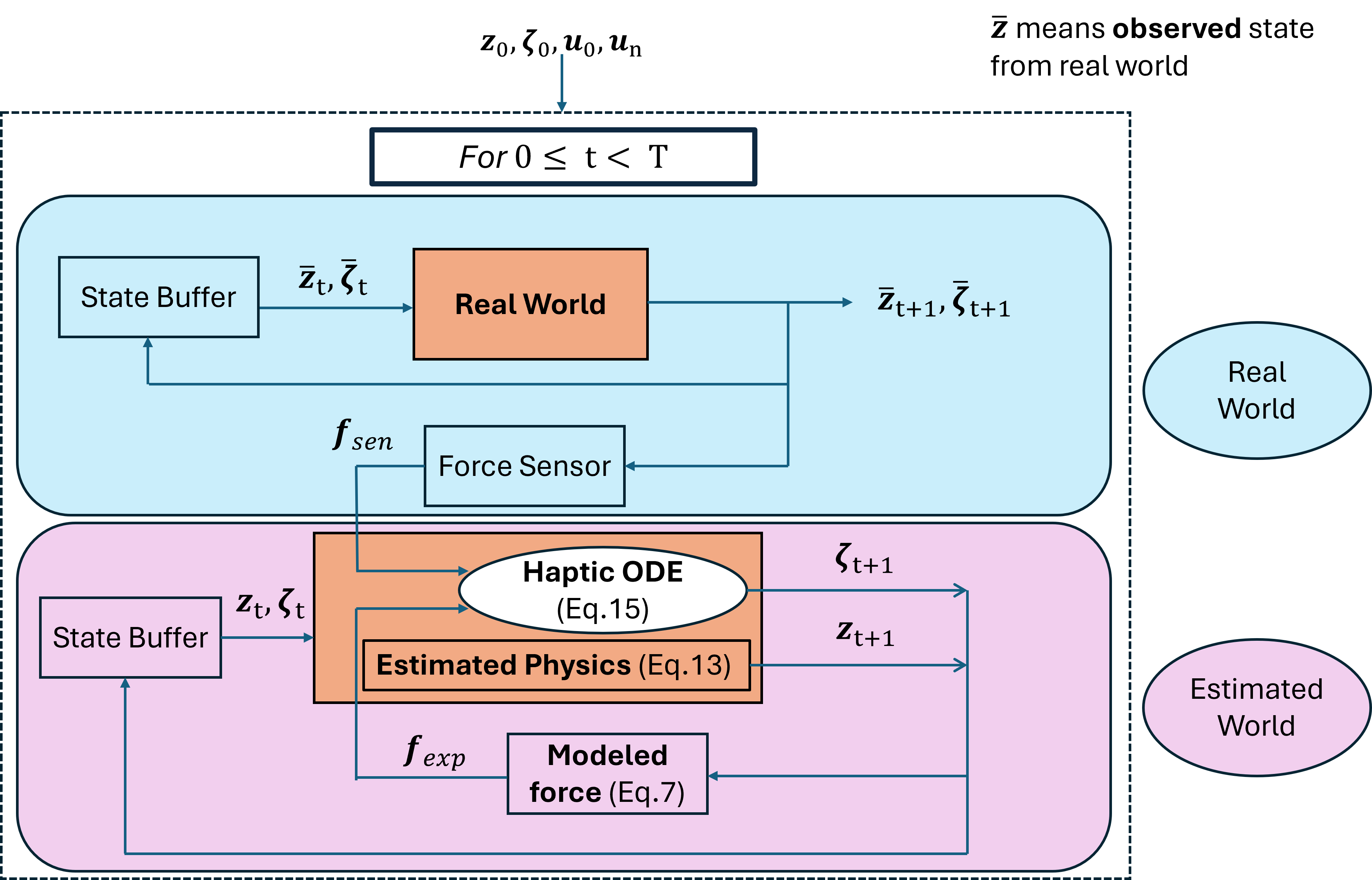}
    \caption{A detailed block diagram demonstrating the concept of the online closed-loop system constituting of Haptic ODE and real world tactile feedback}
    \label{HapticODE}
\end{figure}
\vspace{-2mm}

\section{Experimental Validation}
\begin{figure}[htbp] 
    \centering    \includegraphics[width=0.9\linewidth]{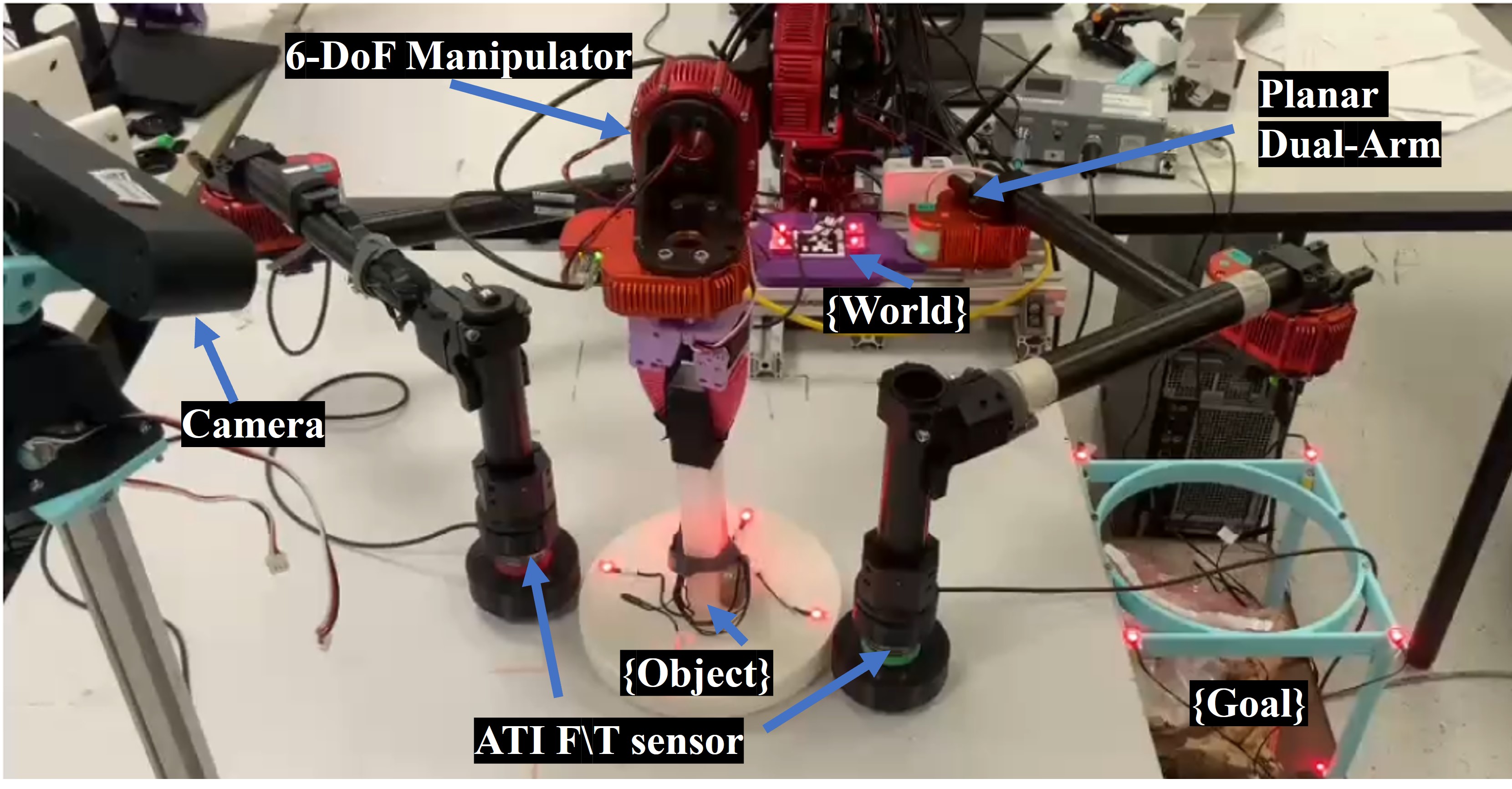}
    \caption{Multi-arm setup, labeled with world, object and goal frame.}
    \label{fig:exp_setup}
    \vspace{-3mm}
\end{figure}

\textbf{Setup}: We carry out both the simulation and the real-world haptic exploration, as well as the sorting task on a smooth table with a multi-arm setup in Fig.\ref{fig:exp_setup}. In simulation, the underlying physics is approximated by iteratively solving the \textbf{Adaptive ODE}. We set $N_{basis} = 20, N_{width}=4, \tau_0 =10$, a zero forcing term for DMP.
In the real world, HEBI X-Series actuators are used to construct the multi-arm setup, while the end-effectors of the planar dual arm are rigidly attached to ATI F/T sensors that measure contact forces. We utilize a ZED2 camera to provide a rough object pose for launching haptic exploration. Ground-truth is collected with PhaseSpace, tracking LED markers on the object. Our scope includes shapes that can be approximated by a single SQ: circular ($\phi$180mm), rectangular (150mm$\times$250mm), elliptical (150mm$\times$250mm), as well as a hexagonal object which is typically constructed by 3 identical rectangles ($L = 150mm, D = 2L\cos{\frac{\pi}{6}}$). All scripts are executed in MATLAB 2024b and we select ODE45 as our solver due to its numerical robustness. 

% First part: haptic exploration
\subsection{Haptic Exploration}
\textbf{Evaluation Metric \& Baseline}: To evaluate the efficacy of the proposed haptic exploration, we compare our shape regression results against a vision-based baseline utilizing Superellipse Recovery from point cloud data \cite{liu2022robust}. Root Mean Square Error (RMSE) is adapted as our evaluation metric for length and width (i.e., $RMSE_L$ and $RMSE_D$) of the regressed shape. We heuristically choose $N=48$ to divide the robot's observation space for our BO-guided haptic exploration algorithm.

% 2x2 Figure for EI/GP progression
\vspace{-2mm}
\begin{figure}[htbp]
    \centering
    \includegraphics[width=0.9\linewidth]{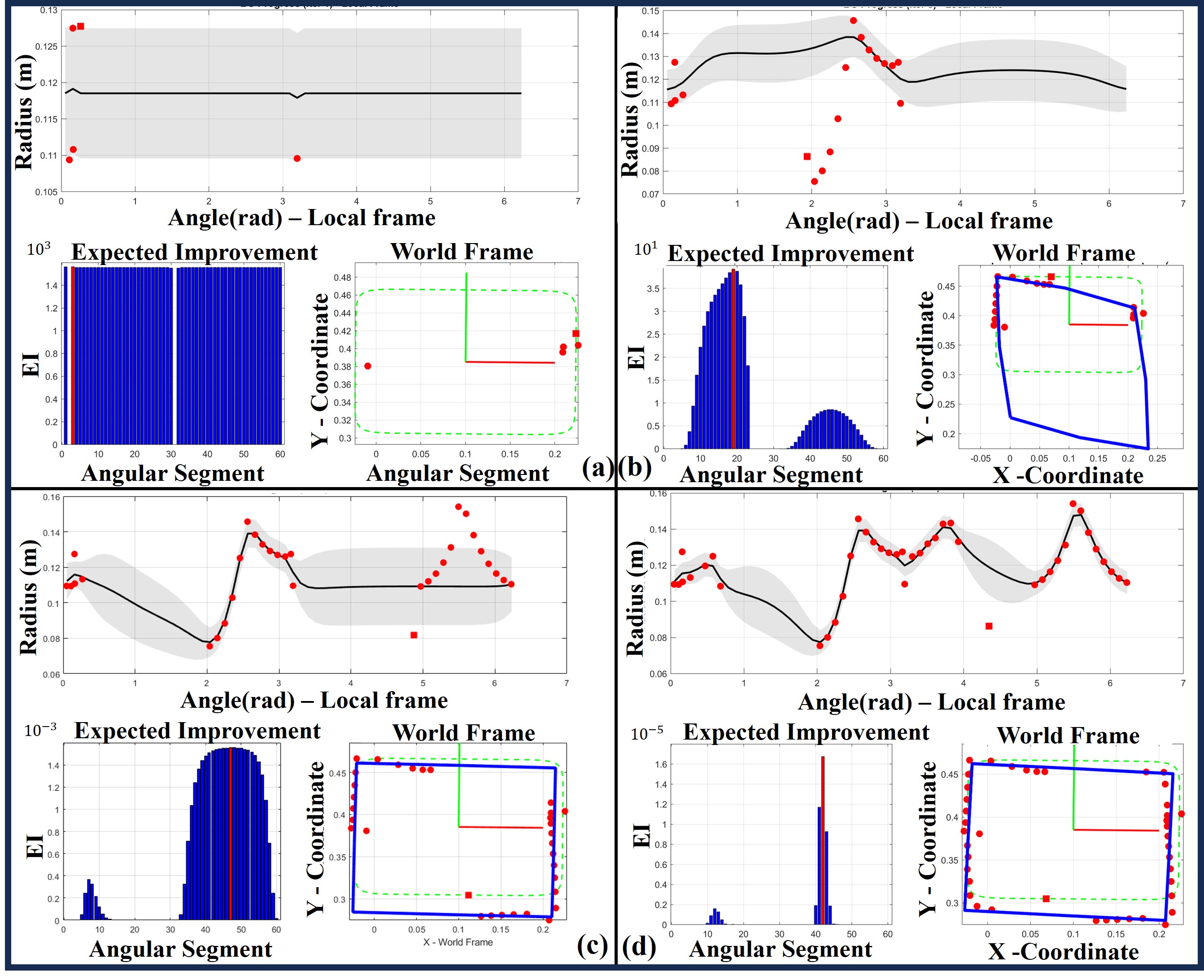}
    \caption{Progression of BO-guided haptic exploration for a rectangular object. We show representative frames: (a) Initial Configuration; (b) After exploring 1 corner; (c) After exploring 2 corners; (d) Algorithm terminates upon convergence.}
    \label{fig:EI_progression}
\end{figure}
\vspace{-2mm}

\textbf{Results}: The progression of the BO-guided haptic exploration for a rectangular object is illustrated in Fig.~\ref{fig:EI_progression}. \HL{Firstly, each robotic agent approaches $\hat\zeta_{cam}$ from two orthogonal directions $(\theta_0, \theta_0+\frac{\pi}{2})$, with $\theta_0$ uniformly sampled from the N=48 segments in each trial.} Starting from an initial configuration with only two contacts, the predictive variance (gray region) reflects a high geometric uncertainty. As the training set is augmented with additional contact points (indicated by red dots), the EI acquisition function proactively directs the robot toward high-uncertainty regions, effectively balancing the exploration of unknown space with the exploitation of identified local features.

% A figure here comparing regressed superellipse from Haptic exploration and point. I am thinking it in form of 2*3 figures
\begin{figure}[htbp] 
    \centering
    \includegraphics[width=0.9\linewidth]{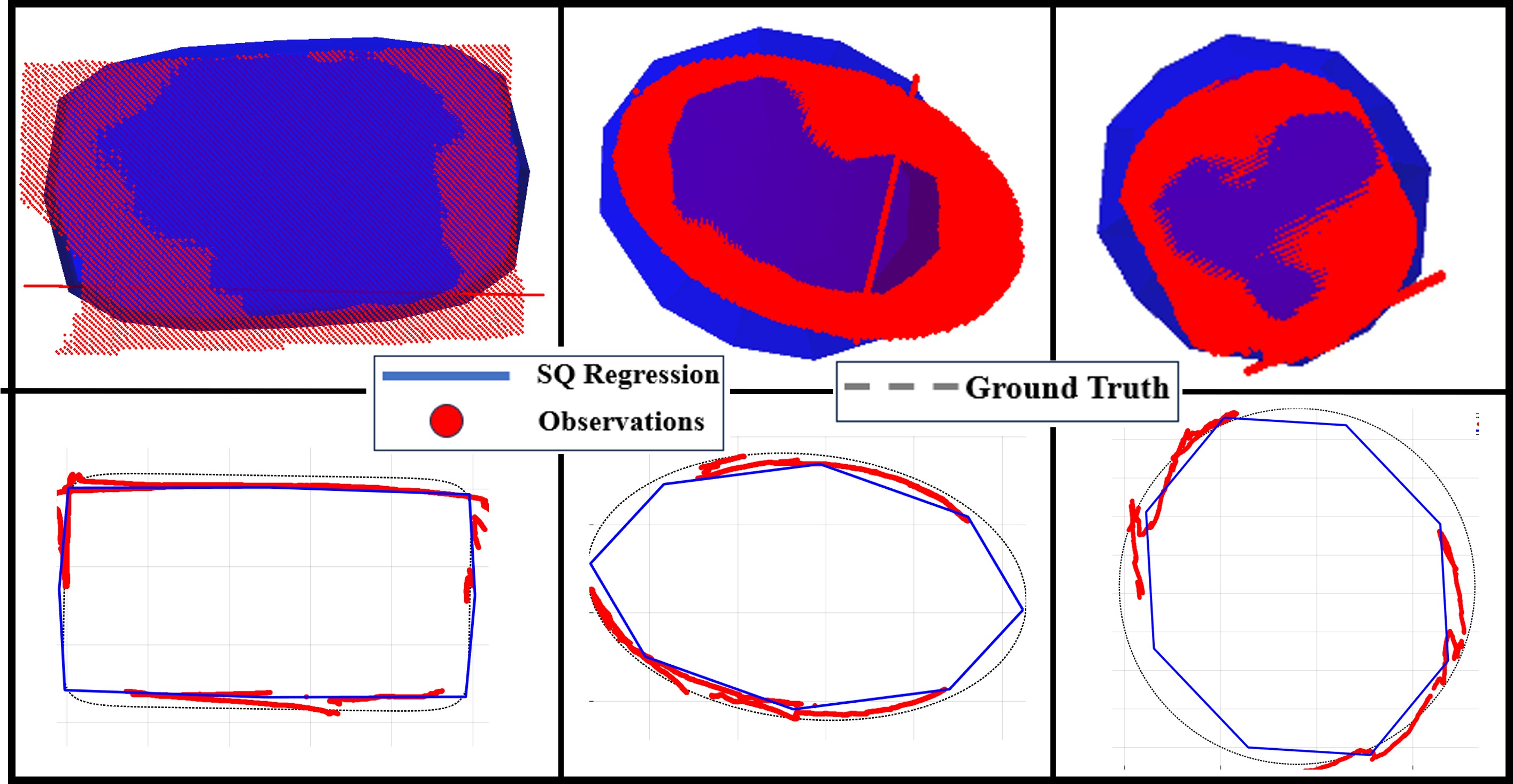}
    \caption{Superquadrics Recovery: (up) Depth-Based Method \cite{liu2022robust}; (down) Haptic exploration ($N = 48$).  The red dots are observations and the blue shape represents the regressed superellipse.}
    \label{fig:BO_result} 
    \vspace{-5mm}
\end{figure}

% Fig. \ref{fig:BO_result} provides a representative visualization comparing our method and the depth-based approach by Liu \textit{et al.} \cite{liu2022robust}. To evaluate the efficiency of the sampling strategy, we introduce \textit{exploration coverage} (see Cover. in Table. \ref{table:exploration_strategy}), formally defined as the ratio of the boundary arc length explored to the total perimeter of the ground-truth geometry. Our method successfully recovers the superquadric geometry without requiring holistic surface exploration, a notable advantage over prior haptic approaches \cite{ottenhaus2018active, suresh2021tactile}. Furthermore, the proposed framework preserves critical geometric features, such as sharp edges, while demonstrating robustness to outliers.
Fig. \ref{fig:BO_result} provides a representative comparison between our method and the depth-based approach by Liu \textit{et al.} \cite{liu2022robust}. To evaluate sampling efficiency, we introduce \textit{exploration coverage} (see Cover. in Table. \ref{table:exploration_strategy}), defined as the ratio of the explored boundary arc length to the total perimeter of the ground-truth geometry. At comparable reconstruction accuracy, lower coverage indicates more efficient exploration. Our method successfully recovers the superquadric geometry without requiring holistic surface exploration, an advantage over prior haptic approaches \cite{ottenhaus2018active, suresh2021tactile}. Furthermore, the proposed framework preserves critical geometric features, such as sharp edges, while demonstrating robustness to outliers.

\vspace{-2mm}
\begin{table}[htbp]
\footnotesize % smaller table
\centering
\caption{\HL{Comparison of haptic strategies in simulation ($\sigma_{thresh}^{sim} = 1mm$) for 10 trials, with different initial contacts, same model, stopping rule and maximum budgets. RMSE values  in ($\mu \pm \sigma$) mm.}}
\label{table:exploration_strategy}
\renewcommand{\arraystretch}{0.95} % smaller column distance
\setlength{\tabcolsep}{4pt}

\begin{tabular}{llccccc}
\toprule
\textbf{Shape} &
\textbf{Metric} &
\textbf{Random} &
\textbf{Variance} &
\textbf{Thompson} &
\textbf{EI (Ours)}
\\
\midrule

\multirow{4}{*}{Circ.}
& Iter. 
& 9.1$\pm$1.4 & 5.6$\pm$0.8 & 6.0$\pm$0.9 & \textbf{5.3$\pm$0.5}\\
& $RMSE_R$
& 0.71$\pm$0.16 & 0.45$\pm$0.07 & 0.43$\pm$0.04 & \textbf{0.39$\pm$0.01}\\
& Cover.\%
& 70.4$\pm$2.7 & 65.3$\pm$1.9  & 64.9$\pm$2.2 & \textbf{64.2$\pm$1.6}\\

\cmidrule(lr){1-6}

\multirow{5}{*}{Rect.}
& Iter. 
& 12.9$\pm1.7$ & 8.6$\pm$0.5 & 7.8$\pm$1.1 & \textbf{6.8$\pm$0.6}\\
& $RMSE_L$
& 0.82$\pm$0.13 & 0.47$\pm$0.04 & 0.47$\pm$0.03 & \textbf{0.45$\pm$0.01}\\
& $RMSE_D$
& 0.88$\pm$0.17 & 0.50$\pm$0.03 & 0.51$\pm$0.04 & \textbf{0.48$\pm$0.02}\\
& Cover.\%
& 80.1$\pm$2.8 & 76.2$\pm$1.6 & 75.0$\pm$2.1 & \textbf{73.5$\pm$1.1}\\

\cmidrule(lr){1-6}

\multirow{5}{*}{Ellip.}
& Iter. 
& 9.4$\pm$1.6 & 7.9$\pm$0.6 & 7.5$\pm$1.0 & \textbf{6.6$\pm$0.8}\\
& $RMSE_L$
& 0.78$\pm$0.15 & 0.44$\pm$0.02 & 0.43$\pm$0.05 &  \textbf{0.41$\pm$0.01}\\
& $RMSE_D$
& 0.83$\pm$0.14 & 0.48$\pm$0.03 & 0.49$\pm$0.03 & \textbf{0.43$\pm$0.01}\\
& Cover.\%
& 79.2$\pm$2.5 & 74.7$\pm$1.1 & 74.0$\pm$1.9 & \textbf{71.6$\pm$1.1}\\

\cmidrule(lr){1-6}

\multirow{5}{*}{Screw}
& Iter.
& 13.1$\pm$2.0 & 8.9$\pm$0.5 & 8.3$\pm$1.1 & \textbf{7.1$\pm$0.5}\\
& $RMSE_L$
& 0.85$\pm$0.15 & 0.53$\pm$0.02 & 0.52$\pm$0.04 & \textbf{0.48$\pm$0.02}\\
& $RMSE_D$
& 0.91$\pm$0.18 & 0.57$\pm$0.06 & 0.55$\pm$0.06 & \textbf{0.50$\pm$0.01}\\
& Cover.\%
& 84.2$\pm$3.0 & 81.1$\pm$1.8 & 80.4$\pm$2.3 & \textbf{77.1$\pm$1.4}\\
\bottomrule
\end{tabular}
\end{table}
\vspace{-2mm}

% Accuracies for x,y,theta, probably 3*3 and place at the top of page 7 ?
\begin{figure*}[!t]
    \centering
    \includegraphics[width=0.99\textwidth]{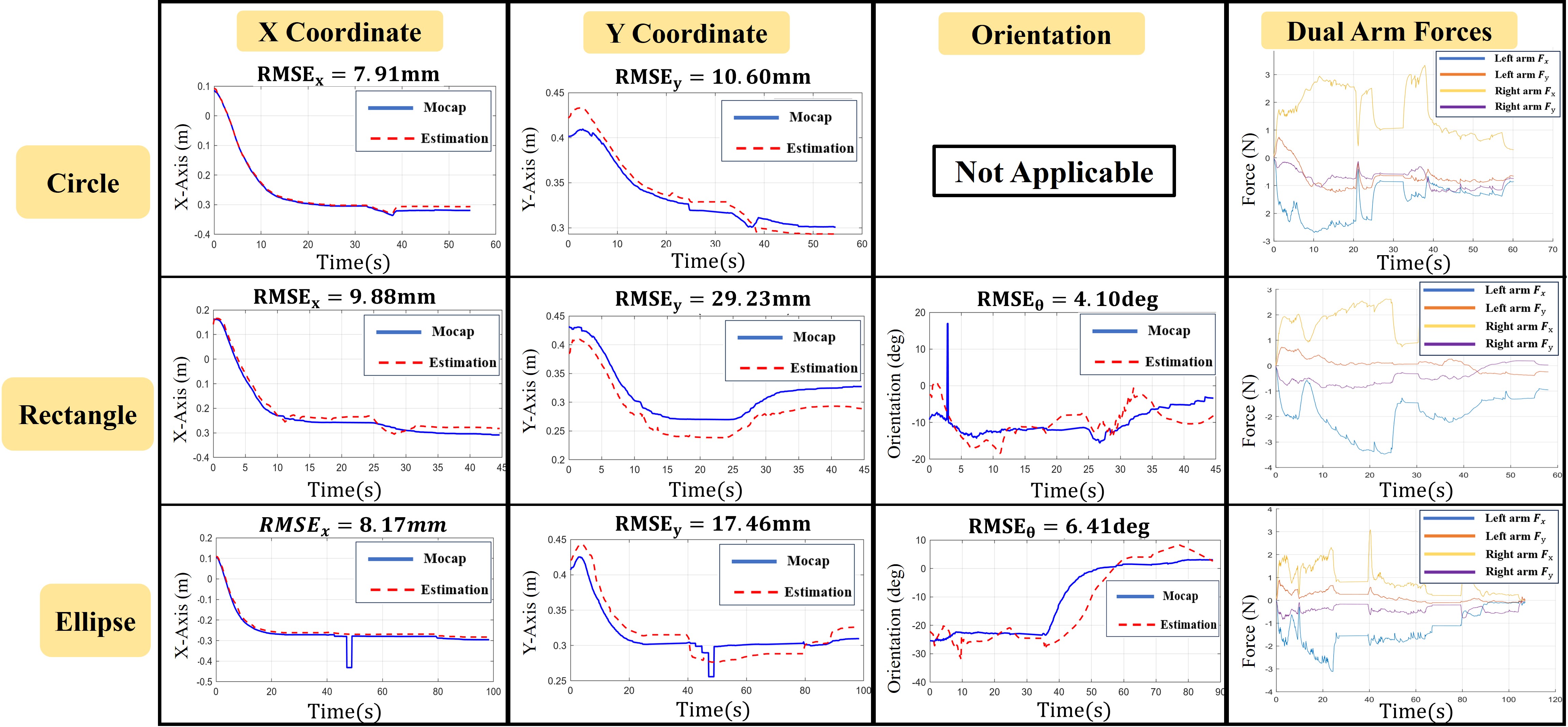}
    \caption{Representative results of the \textbf{average} pose estimation accuracy (measured in Per-Axis RMSE) against motion capture, alongside with respective force profiles. The red dotted line is the Haptic ODE estimation while the blue solid line shows the ground truth from motion capture system. The unmeant spikes in motion capture recording are mainly due to occlusion of LED markers during the experiments.}
    \label{fig:estimation_accuracy}
    \vspace{-3mm}
\end{figure*}

In Table \ref{table:exploration_strategy}, we compare our framework (EI) against:
\begin{enumerate}
    \item Random: randomly select an angle segment to explore.
    \item Variance: select the angle segment with the largest posterior variance of the GP.
    \item Thompson: samples from the joint posterior distribution of the GP.
\end{enumerate}
The comparative analysis validates the geometric accuracy and exploration efficiency of our framework against alternative tactile sampling policies in simulation. Uncertainty-driven acquisition (e.g., Maximum Posterior Variance/Entropy) tends to waste contacts on regions with high predictive variance but low geometric significance. Furthermore, while sampling-based methods like Thompson Sampling naturally balance exploration and exploitation, they introduce a degree of stochasticity into the robot action.

\begin{table}[htbp]
\centering
\caption{Real-world comparison between point-cloud-based perception and haptic exploration ($\sigma_{thresh}^{real}=5$ mm) over 10 trials. RMSE values  in ($\mu \pm \sigma$) mm.}
\label{table:real_comparison}
\renewcommand{\arraystretch}{0.95}

\begin{tabular}{llcc}
\toprule
\textbf{Shape} &
\textbf{Metric} &
\textbf{Point Cloud \cite{liu2022robust}} &
\textbf{Haptic (Ours)}
\\
\midrule
\multirow{4}{*}{Circular}
& $RMSE_R$
&16.38$\pm$1.01
&\textbf{1.96$\pm$0.12}
\\
& BO Iterations
&N.A.
&\textbf{5.6$\pm$0.5}
\\
& Expl. Coverage
&N.A.
&\textbf{65.1$\pm$1.2\%}
\\

\cmidrule(lr){1-4}
\multirow{5}{*}{Rectangular}
& $RMSE_L$
&22.05$\pm$1.47
&\textbf{3.81$\pm$0.15}
\\
& $RMSE_D$
&18.78$\pm$1.29
&\textbf{5.77$\pm$0.13}
\\
& BO Iterations
&N.A.
&\textbf{7.0$\pm$0.8}
\\
& Expl. Coverage
&N.A.
&\textbf{80.9$\pm$1.6\%}
\\

\cmidrule(lr){1-4}
\multirow{5}{*}{Elliptical}
& $RMSE_L$
&18.31$\pm$1.89
&\textbf{3.76$\pm$0.18}
\\
& $RMSE_D$
&13.25$\pm$0.83
&\textbf{4.89$\pm$0.14}
\\
& BO Iterations
&N.A.
&\textbf{6.7$\pm$0.6}
\\
& Expl. Coverage
&N.A.
&\textbf{78.2$\pm$1.3\%}
\\
\bottomrule
\vspace{-3mm}
\end{tabular}
\vspace{-5mm}
\end{table}

As shown in Table \ref{table:real_comparison}, our method achieves millimeter-level accuracy across all tested geometries, outperforming the depth-based baseline \cite{liu2022robust} which exhibits errors extending into the centimeter range. The performance gap highlights the fundamental limitations of vision-based recovery in unstructured settings, as the recovery algorithm extrapolates holistic superquadric parameters from a degraded, partial observation. \HL{A two-way ANOVA (Shape $\times$ Strategy) on simulation RMSE showed significant main effects of Shape ($F_{3,144}=12.18$) and Strategy ($F_{3,144}=176.80$; both $p<0.001$) without interaction ($F_{9,144}=0.51$, $p=0.868$), i.e., the relative performance of strategies is consistent across geometries. Tukey HSD tests showed that EI significantly outperforms Random ($p<0.001$), whereas its differences from Variance ($p=0.113$) and Thompson ($p=0.335$) are not significant. On real-world data, Shape ($F_{2,54}=62.17$), Method ($F_{1,54}=3266.86$) and their interaction ($F_{2,54}=20.69$) were all significant ($p<0.001$). Haptic exploration reduced RMSE by 15.74\,mm on average ($p<0.001$), and per-shape paired $t$-tests confirmed a significant improvement for every shape.}

% Second part: robotic sorting
\subsection{Robotic Sorting}

% talk about the goal and initial pose. 
\textbf{Setup}: We carry out a sorting task immediately after the termination of haptic exploration and the computation of manipulation potential. The Haptic ODE first estimates an initial object pose $\hat\zeta_0$ based on tactile feedback and runs in parallel to plan sorting trajectories toward the goal pose $\zeta_{goal}$. \HL{The task is considered a "Success" if the final estimated pose \( \hat{\zeta}_t = [\hat{x}_t, \hat{y}_t, \hat{\theta}_t]^T \) reaches the target \( \zeta_{goal} \) with a per-axis error of less than \( 10 \) mm and an absolute orientation error of less than \( 3^\circ \) (position only for the circle).}

% figure demonstrating the entire sorting process: 2*2
% \begin{figure}[h!] 
% \vspace{-2mm}
%     \centering
%     \includegraphics[width=0.8\linewidth]{figures/sorting_v2.jpg}
%     \caption{Sorting a Rectangular object: (a) Initial configuration; (b)-(c) Adaptive trajectory re-planning; (d) Task Completion.}
%     \label{fig:sorting}
%     \vspace{-2mm}
% \end{figure}

\begin{figure}[!ht] 
    \centering
    \includegraphics[width=\linewidth]{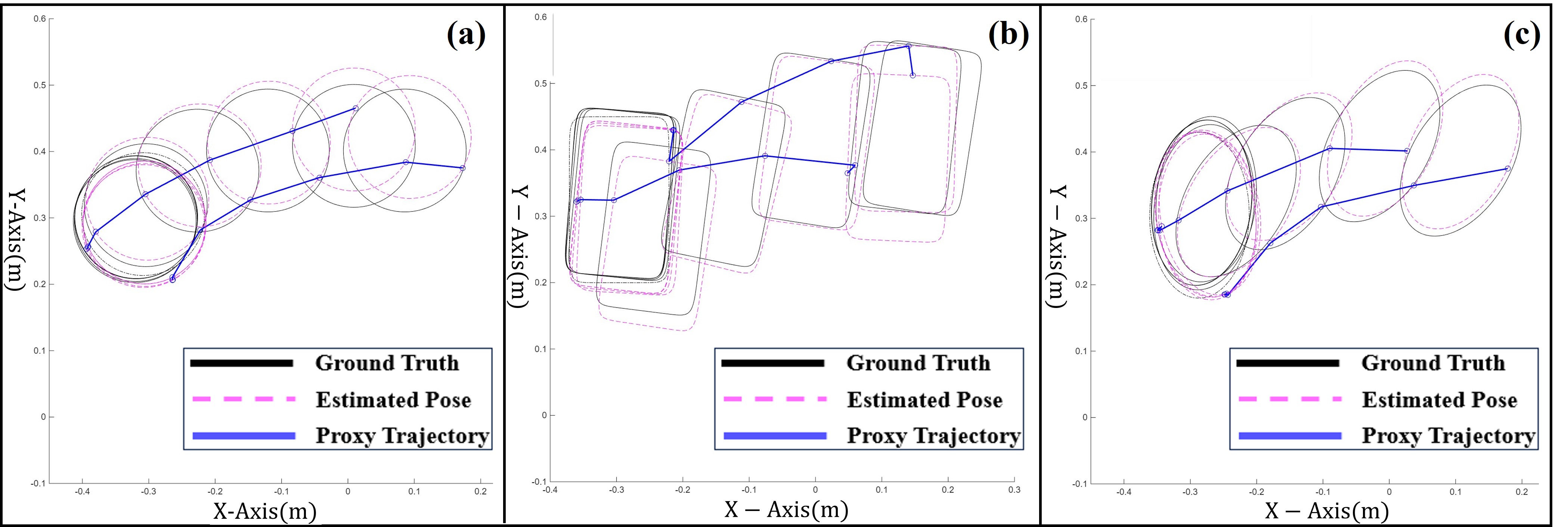}
    \caption{Sorting trajectories: (a) circular; (b) Rectangular; (c) Elliptical. The black, pink geometries represent the ground truth and estimation respectively, while the blue dots denote contacts.}
    \label{fig:traj_overview}
    \vspace{-3mm}
\end{figure}

\textbf{Results}: Fig.\ref{fig:traj_overview} visualizes the ground truth trajectory alongside the \textbf{Haptic ODE} estimations of (a) circular, (b) rectangular, and (c) elliptical objects. Due to the imperfect grasp and physical disturbances inherent in real-world interactions, the object inevitably deviates from the intended path. To counteract this, our online ODE provides asynchronous correction to the sorting trajectory.

\vspace{-2mm}
% estimation against Mocap: 2*2

We evaluate the pose estimation accuracy of the proposed Haptic ODE framework by comparing the estimated trajectories against a motion capture system, which provides the absolute ground-truth object pose, $\bar{\zeta}$.
Fig. \ref{fig:estimation_accuracy} presents a quantitative analysis of the object's planar position ($x, y$) and orientation ($\theta$) over the duration of the manipulation task. 
The framework demonstrates robust translational tracking over all 10 trials per geometry. The overall per-axis positional RMSE remains bounded below $30$ mm, and finally converges to the goal pose at the end of the trajectory. For objects possessing distinct geometric features (i.e., rectangular and elliptical profiles), the orientational RMSE is maintained at approximately $5^{\circ}$, indicating stable rotational inference. Lastly, the force profiles reveal the closed-loop correction mechanism of the Haptic ODE. When unmodeled physical perturbations (i.e., grasp slip) cause the object to deviate from its nominal path, the resulting contact forces naturally escalate. After 2 or 3 active corrections, the sensed forces stabilize as the object approaches the target pose.

\textbf{Benchmark and Ablation Study:} We conduct a comparative benchmark and an ablation study on a 2D multi-arm sorting task (Table \ref{table:sorting_result}). The evaluation comprises three distinct conditions: (1) our Haptic ODE with shape parameters regressed from haptic exploration; (2) our Haptic ODE with superellipse parameters regressed from the point cloud \cite{liu2022robust}; and (3) an ablation study without the Haptic ODE, relying entirely on a 6-DoF open-loop position controller.

As shown in Table \ref{table:sorting_result}, our framework (Haptic Expl. + Haptic ODE) exhibits a noticeable performance advantage, achieving high success rates across all object geometries. The average cycle time of the Haptic ODE solver is approximately 3$ms$, ensuring a high observer frequency of about 300$Hz$. The ablation study (w/o Haptic ODE) demonstrates the baseline difficulty of an open-loop controller, as it lacks the adaptability for physical uncertainties.

Notably, the Haptic ODE with point-cloud derived parameters performs \emph{worse} than the open-loop ablation across all three geometries: biased shape parameters produce systematically wrong force predictions, and the closed-loop correction drives the pose estimate in incorrect directions. This counter-intuitive result quantifies the criticality of shape accuracy in our gradient-based observer, which relies on the predicted force ${\mathbf{f}}_{exp}=-{\partial }_{\mathbf{u}}W\left({\mathbf{z}}^*(\mathbf{u}, \V \zeta),\mathbf{u},\V \zeta \right)$ and the gradient 
$\mathbf{\nabla }F\left({\mathbf{c}}_{ij}^{\ast }\right)$ 
computed based on the superellipse geometry.

\vspace{-2mm}
\begin{table}[H] 
\centering
\caption{Sorting performance over 10 trials per shape. Haptic columns report success count and mean 2D trajectory error (mm); the open-loop baseline reports success count.}
\label{table:sorting_result}
\setlength{\tabcolsep}{4pt}
\begin{tabular}{lccc} 
\toprule
 & \multicolumn{2}{c}{\textbf{With Haptic ODE}} & \multirow{2}{*}{\textbf{w/o Haptic}} \\
\cmidrule(lr){2-3}
\textbf{Shape Profile} & \textit{Ours} & \textit{Point Cloud \cite{liu2022robust}} & \\ 
\midrule
Circular    & \textbf{10/10(14.57$\pm$1.49)} & 3/10(39.42$\pm$1.82) & 6/10 \\ 
Rectangular & \textbf{10/10(28.71$\pm$2.06)} & 1/10(49.74$\pm$2.35) & 2/10 \\
Elliptical  & \textbf{8/10(19.43$\pm$1.75)}  & 1/10(45.96$\pm$2.53) & 2/10 \\
\bottomrule
\end{tabular}
\end{table}
\vspace{-4mm}

\section{CONCLUSIONS}
We propose a unified, manipulation-centric framework in which SQ parameterization simultaneously supports haptic exploration, modeling, and an online haptic-driven pose observer. Our framework enables a multi-arm robotic system to autonomously explore, model, and manipulate objects with unknown initial geometries and poses.
Our BO-guided haptic exploration successfully recovers SQ parameters without exhaustive surface coverage, and we empirically demonstrate that local haptic feedback is sufficient to track object states. Ablation reveals that the Haptic ODE is highly sensitive to shape accuracy. This quantitatively validates the necessity of a unified perception-to-control pipeline rather than treating shape estimation as a decoupled preprocessing step.

\HL{Our current scope is limited to convex objects (i.e. circle, rectangle, ellipse and hexagon) and 2D interactions in uncluttered scenes. Future work will consider alternative geometric distributions, and extend the proposed framework into a unified visuo-tactile system that leverages the complementary strengths of both perceptual modalities.}

%%%%%%%%%%
%Reference
%%%%%%%%%%
\bibliographystyle{ieeetr}
\bibliography{myref}

\end{document}